\documentclass{article}
\usepackage{ijcai22}

\usepackage{times}
\usepackage{soul}
\usepackage{url}
\usepackage[hidelinks]{hyperref}
\usepackage[utf8]{inputenc}
\usepackage[small]{caption}
\usepackage{graphicx}
\usepackage{amsmath}
\usepackage{amssymb}
\usepackage{amsthm}
\usepackage{booktabs}
\usepackage{xcolor}
\usepackage{multirow}
\usepackage{multicol}
\usepackage{float}
\usepackage{tabularx}
\usepackage{ifthen}

\usepackage[ruled, linesnumbered]{algorithm2e}
\usepackage{makecell}

\title{Prompting to Distill: \\
Boosting Data-Free Knowledge Distillation via Reinforced Prompt}

\author{
Xinyin Ma$^1$
\and
Xinchao Wang$^2$\and
Gongfan Fang$^1$\and
Yongliang Shen$^1$\And
Weiming Lu$^1$\thanks{Corresponding author}
\affiliations
$^1$College of Computer Science and Technology, Zhejiang University \\ 
$^2$Department of Electrical and Computer Engineering, National University of Singapore
\emails
\{maxinyin,fgf,syl,luwm\}@zju.edu.cn,
xinchao@nus.edu.sg
}

\providecommand{\modelname}{PromptDFD}
\providecommand{\pseudoG}{\ensuremath{\mathcal{G}_{c}}}
\providecommand{\promptG}{\ensuremath{\mathcal{G}_{p}}}
\providecommand{\pr}{\ensuremath{\operatorname{Pr}}}
\providecommand{\expect}{\ensuremath{\mathbb{E}}}
\newcommand{\AppendixFlag}{true}

\begin{document}

\maketitle

\begin{abstract}


Data-free knowledge distillation (DFKD) conducts 
knowledge distillation 
via eliminating the dependence of
original training data,
and has recently 
achieved impressive results in accelerating pre-trained language models. 
At the heart of DFKD is to
reconstruct a synthetic dataset by inverting
the parameters of the uncompressed model.
Prior DFKD approaches, however, have
largely relied on 
hand-crafted priors of the target data distribution
for the reconstruction, 
which can be inevitably biased
and often incompetent to capture the 
intrinsic distributions.
To address this problem, we propose a prompt-based method, termed as PromptDFD,
that allows us to take advantage of learned language priors,
which effectively harmonizes the synthetic sentences to be semantically and grammatically correct.
Specifically, 
PromptDFD leverages a pre-trained generative model to provide language priors and introduces a reinforced topic prompter to control data synthesis, making the generated samples thematically relevant and 
semantically plausible, and thus friendly to downstream tasks. 
As shown in our experiments, the proposed method substantially 
improves the synthesis quality and  
achieves considerable improvements on distillation performance.
In some cases, PromptDFD
even gives rise to results 
on par with 
those from the data-driven knowledge distillation 
with access to the original training data.

\end{abstract}

\section{Introduction}

\begin{table}[h]
    \small
    \centering
    \scalebox{0.93}{
        \begin{tabular}{p{1.6cm}p{6.7cm}}
        \midrule \midrule
        Method & Synthesized Sample\\
        \midrule
        AS-DFD & ... indonesia (logctricdae beings eantarize whilellus civilian indiana influencesidium landed gaddafi ...\\
        \modelname\ & \emph{\textbf{[Topic Prompt]}} If the obama administration is \emph{\textbf{[Content]}} planning to release all of Obama's ``red-book" travel documents, but it's unclear if they include information on Obama's first foreign trip. The documents, as leaked yesterday by ABC News ...\\
        \midrule \midrule
        \end{tabular} 
    }
    \caption{Examples of pseudo samples generated for the category ``World'' in AG News. We use nearest neighbor search to approximate the tokens for synthesized embeddings generated by AS-DFD to make the synthesized embeddings readable.
    \label{tbl:sample_compare}}
\end{table}
        

Knowledge distillation (KD) has recently 
emerged as a popular 
technique for model lightening,
finding its applications in a wide spectrum of domains.
In some scenarios, however, the training data {per se}
is unavailable, which calls for 
a variant of KD, known as
\emph{data-free knowledge distillation}~(DFKD).
Despite the more challenging problem setting,
DFKD has also received increasing attention from the community
due to its practical nature~\cite{micaelli2019zerozskt,yin2019dreaming,Ma2020ASDFD,Fang2021CMI}. 

In the literature, \cite{lopes2017data} initially introduces the idea of data-free knowledge distillation, and follows a learning-by-generating paradigm to craft synthetic data that approximates the original distribution. The key step towards DFKD is to generate high-quality and diverse pseudo samples from a distribution similar to that of the original training set. 
Once the pseudo samples have been generated, conventional KD algorithms \cite{Sun2019PatientKD,Jiao2020Tinybert,Hou2020DynaBERT} can be easily deployed to transfer knowledge from the teacher to the student network. 

Prior DFKD algorithms in natural language processing \cite{Ma2020ASDFD,Rashid2021zskdnlp}
focused on synthesizing pseudo samples from the teacher's parameters through model inversion \cite{mahendran2015understanding}, where a batch of synthetic utterances or a sentence generator is optimized under the restrictions of some human-crafted distributional priors. 
The confidence-based prior is the most widely used human-crafted prior for sentence synthesizing. For example, AS-DFD \cite{Ma2020ASDFD} aims to find some pseudo samples that can produce high-confidence predictions when fed to the teacher. 
As shown in Table \ref{tbl:sample_compare}, despite that AS-DFD indeed generates some task-related keywords or phrases that are related to the task, these utterances are still unnatural and of low-quality without correct semantic and syntax. 

The above issue is in part caused by the insufficiency of the confidence-based prior in regularizing the synthesized samples. For classification problems, the teacher's predictions mainly contain high-level information like the category of sentences, from which the low-level information  (correlation between words, syntax, etc.) is removed. Therefore, the confidence-based prior alone cannot provide robust regularization for the low-level features to make the whole sentences plausible, and leads to some out-of-distribution samples known as ``rubbish samples''~\cite{goodfellow2015explaining}. These samples are far from the underlying distribution of the original training set and are hardly helpful for distillation. Existing methods address this problem by introducing more regularizations into data synthesis. For instance, \cite{yin2019dreaming} adds extra restrictions using the stored mean and variance statistics in Batch Normalization to improve the quality of synthesized samples. However, these methods are usually inapplicable for language models as they heavily rely on some special architectures like Batch Normalization that are not used in transformer-based models. Therefore, how to introduce more priors to regularize the synthesized sentence is still a challenging problem for DFKD.



To address these problems, we resort to learning natural priors in pre-trained generative language models, and develop a prompting-to-distill framework for DFKD which we term as \modelname. Generative language models pre-trained on a large scale of unlabeled corpora grasp the generic prior for languages, which can be incorporated into the synthesis to alleviate the quality issue caused by confidence-based prior. 
Specifically, the generation of synthesized samples is divided into two steps: 1) a prompt step that explores the language prior in the pre-trained language model in a controllable manner by a task-related topic prompt, and 2) a completion step to expand the prompt into a complete utterance that contains more informative details. 
To enforce the prompt to be tightly related to the target task when no original data is available, we propose an adversarial reinforced prompt that takes the feedback from both the teacher and the student model into account, simultaneously satisfying the topic and the difficulty requirement. 
As shown in the experiment, \modelname\ significantly improves the synthesis quality and boosts the student's performance. Extensive experiments demonstrated that the synthesized samples generated by \modelname\ not only contain task-related words, but are meaningful and follow grammatical and linguistic conventions. 

In summary, the contribution of this paper can be summarized as follows:

\begin{itemize}
    \item We propose a novel prompting-to-distill framework for data-free knowledge distillation, through introducing controllable language prior into the synthesized sample.
    \item We propose a reinforced topic prompter with an adversarial reward. With no access to the original training set, the reinforced prompter can dynamically adjust the topic and the difficulty of the synthesized sample to satisfy the topic constraint of the target tasks 
    \item \modelname\ achieves superior performance on four text classification datasets with different network settings.
    In some datasets, \modelname\ even yields results on par with those of data-driven KD methods.
\end{itemize}

\section{Related Work} \label{sec:related}
\paragraph{Data-driven knowledge distillation.}
Knowledge distillation (KD), proposed by \cite{Hinton2015Distill}, uses one or more large and cumbersome models (called the teacher model) to train a lightweight model (called the student model), where the student is required to mimic the softened response of the teacher. DistillBERT \cite{Sanh2019DistilBert} was the first work to adopt KD on BERT, after which PatientKD \cite{Sun2019PatientKD}, Tinybert \cite{Jiao2020Tinybert}, Universal-KD \cite{wu2021UniversalKD} and other algorithms have developed other feature alignment criteria, such as intermediate representations, attention maps, extracted relations, etc. Some works \cite{Hou2020DynaBERT,Chen2020DAdaBERT} focuses on how to dynamically select the most valuable structure of student models during knowledge distillation. Those KD algorithms can be applied as alternative ways to distill knowledge when synthesized samples are obtained.

\paragraph{Data-free knowledge distillation.}
Existing DFKD algorithms in computer vision adopt the idea of model inversion \cite{mahendran2015understanding} to reconstruct the training samples from the parameters of the model. Under the constraint of the probabilistic output \cite{nayak2019zerozskd} or historical statistics stored in Batch Normalization \cite{yin2019dreaming}, the training samples \cite{lopes2017data} or pseudo-sample generators \cite{chen2019data} are optimized to explore the underlying distribution of the training data stored in the teacher network. \cite{micaelli2019zerozskt} quantifies the degree of belief matching between the teacher and student. \cite{Fang2021CMI} proposes to solve the problem of inter-sample pattern collapse using contrast learning. As for methods designed for natural language processing, \cite{Ma2020ASDFD} generates pseudo-embeddings to address the discrete problem of tokens, and \cite{Rashid2021zskdnlp} adopts Gumbel-Softmax to pass the gradient to the generator. However, the only constraint that is accessible in BERT is the class prior, which is too weak to reconstruct the underlying distribution.

\begin{figure*}[t]
    \centering
    \scalebox{0.44}{
    \includegraphics{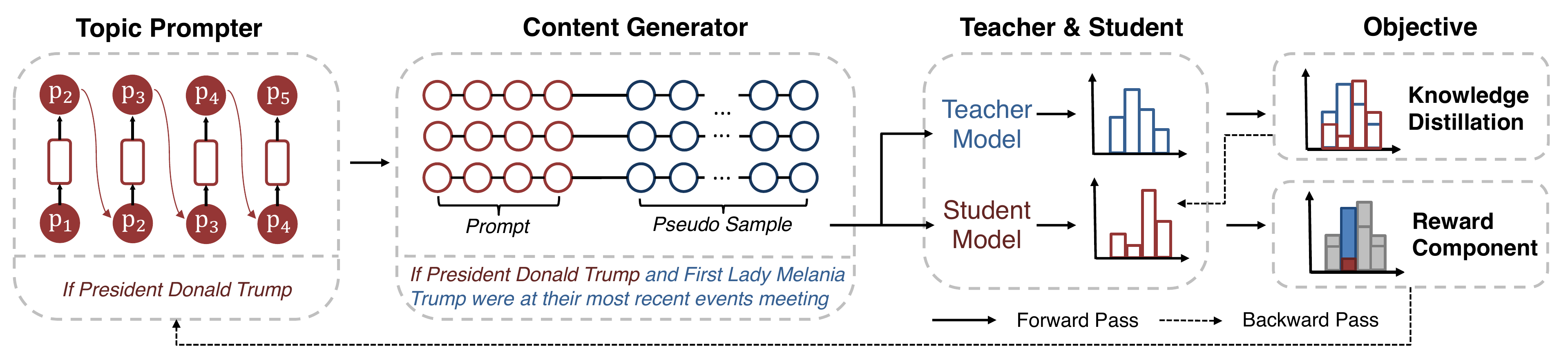}
    }
    \caption{The illustration of the proposed \modelname. Synthesizing a sample comprises two steps: generating a topic prompt and expanding the prompt into a full sentence. The synthesized utterance is then used to train the student model with a conventional KD objective and adjust the generation of the topic prompter with an adversarial reward component. }
    \label{fig:my_model}
\end{figure*}

\section{Problem Settings} \label{sec:preliminary}
The goal of data-free knowledge distillation is to reduce redundant parameters in the network without relying on the original dataset. We first start from the data-driven scenario. 

\paragraph{Data-driven knowledge distillation.}
Model compression can be achieved using knowledge distillation, in which the original model is called the teacher network $\mathcal{T}$ and the compressed model is called the student network $\mathcal{S}$. Given a sample $x = [t_1, t_2, \dots, t_l] \in \mathcal{X}$ and its pre-defined label $y$, where $t_i$ is the i-th token in the sample and $l$ is the length of the sentence, the objective of KD can be defined as follows:
\begin{equation}
    \mathcal{L}_{KD} = \alpha \cdot \mathcal{L}_{CE}(\mathcal{S}\left(x\right), y) + (1-\alpha) \mathcal{L}_{KL}(\mathcal{T}\left(x\right), \mathcal{S}\left(x\right)) \label{formula:KD_loss}
\end{equation}
where $\mathcal{T}(x)$ and $\mathcal{S}(x)$ are the predicted class probabilities on input $x$. $\alpha$ balances between two criteria, i.e., Cross-Entropy loss $\mathcal{L}_{CE}$ and Kullback-Leibler divergence loss $\mathcal{L}_{KL}$.

\paragraph{Data-free knowledge distillation.} 
Data-free knowledge distillation is severely hampered by the lack of access to the initial distribution of data $\mathcal{X}$, and as a consequence, supervisory signals and predefined labels are not available. Thus, constructing the synthesized sample $\hat{x} \in \hat{\mathcal{X}}$ that can reflect the distribution of $\mathcal{X}$ is the key to the data-free knowledge distillation. After the transfer set $\hat{\mathcal{X}}$ is rebuilt, knowledge distillation techniques can be applied to minimize $\mathcal{L}_{KD}$.

\section{Method}


\subsection{Prompting to Distill} \label{sec:prefix-prompt}
Model inversion with the one-hot class prior fails to constrain the low-level correlation in context, resulting in the semantic confusion of the utterance. The auto-regressive pre-trained language models, such as GPT2, are trained on large-scale corpora to capture these informative co-occurrences. Inspired by \cite{Shin2020AutoPrompt} and \cite{Li2021prefixtuning}, our intuition is to query a pre-trained generative language model by prompt engineering and explore the language prior of the language model in a controllable manner.

\paragraph{Step 1: Prompt step.}
A prefix prompt is used to restrict the topic of the entire synthesized sentence, ensuring that the synthesized sentences are assigned some topics that are relevant to the original training set. Specifically, the topic prompt $P_{1:n} =( p_{1}, p_{2}, \dots, p_{n} )$, where $n$ is the length of the prompts, contains a shortlist of introductory words and can be considered as the controller for the pre-trained generative LM in step 2. We will discuss the design of topic prompts in section \ref{sec:prompt_design}.

\paragraph{Step 2: Completion step.}
Based on the generated prompt, we adopt a pre-trained auto-regressive generator, named the content generator \pseudoG, to extend the prompt and explicitly display the knowledge embedded in the pre-trained LM. To avoid the knowledge in the pre-trained LM being disturbed when tuning, we freeze the parameter $\theta_{\mathcal{G}}$ of \pseudoG. The goal of \pseudoG\ is to extend the topic prompt $P_{1:n}$ and generate the content $\hat{x}$, which behaves as the synthesized sample in knowledge distillation. Given a half-finished sequence $ (\hat{x}_{1}, \hat{x}_{2}, \dots, \hat{x}_{i})$ generated by \pseudoG, the auto-regressive LM computes the hidden states based on the past tokens and the prompt:
\begin{align}
    & h_{i} = \pseudoG ([p_{1}, \dots, p_{n}, \hat{x}_{1}, \dots, \hat{x}_{i}]; \theta_{\mathcal{G}}) \nonumber \\
    & \pr(\hat{x}_{i+1}|P, \hat{x}_{\leq i}) = \operatorname{softmax}(Wh_{i})
\end{align}
where $h_{i}$ is the hidden state for the temporary token and $W$ projects the hidden state into the distribution space of the tokens. For \pseudoG, we use pre-trained GPT2 \cite{radford2019language} for content filling. When decoding, a top-k nucleus sampling \cite{Holtzman2019Top-P} is used to encourage the diversity of pseudo samples. 

\subsection{Prefix Prompt for Topic Controlling} \label{sec:prompt_design}
\paragraph{Hand-crafted prompt.} We start with a straightforward way of designing a topic prompt by manually selecting some keywords that are pertinent to the task. Table \ref{tbl:manual_prompt} contains a few templates that are designed manually. While hand-crafted prompts are capable of guiding the generation of pseudo-samples within a given subject area, there are two major inherent flaws: it is difficult to determine the optimal way to design prompts, and the number of hand-crafted prompts is limited. Though decoding with sampling can provide some variability and randomness in the generation process, the content of the synthesized sample is restricted to limited information.

\begin{table}[t]
    \centering
    \begin{tabular}{lcc}
    \midrule \midrule
    Datasets & Hand-crafted Prompt  \\
    \midrule
    AG News & A latest [Category] news \\
    DBPedia & A document about [Category] \\
    IMDb & A [Category] movie review \\
    SST-2 & [Category] sentence: \\
    \hline \hline
    \end{tabular} 
    \caption{Manually designed prompts for prefix prompting. [Category] represents the name of the category, such as positive or negative for IMDB, world, sports, business or science for AG News.} \label{tbl:manual_prompt} 
\end{table}

\paragraph{Reinforced prompt.} To solve the problem encountered when designing prompts, we propose to generate prompts automatically. We train a topic prompter \promptG, which can be tuned to generate topic prompts that are closely related to the domain of the original dataset.

The generation of prompt $P_{1:n}$ can also be defined as a left-to-right sequence generation problem. Suppose that $P_{1:t} = (p_{1}, \dots, p_{t})$ has been generated, it's desired to infer the probability distribution of the next token $\pr(p_{t+1}|P_{1:t})$. Unlike the training objective in the sequence generation problem in general, in DFKD there is no immediate and ground-truth label for $\pr(p_{t+1}|P_{1:t})$. Furthermore, the teacher model may take a different tokenization method with the content generator \pseudoG, leading to the mismatch in the subwords. Thus, previous gradient-based methods that synthesize embeddings or leverage the Gumbel-Softmax cannot be applied. 
 
We propose a reinforced prompter, which tunes itself through the feedback from both the teacher and the student model. Suppose that the state $s_t$ for the current moment $t$ is defined as the generated sequence $P_{1:t}$ and the action $a_t$ as the selection of the next token $p_{t+1}$. The selecting policy $\pi$ for the next action is inferred by the prompt generator \promptG with the input being the generated sequence and the output being the probabilities of each token as follows:
\begin{equation}
    a_t \sim \pi \left( s_t, \phi_{\mathcal{G}} \right) = \promptG (p_{t+1}|P_{1:t}, \phi_{\mathcal{G}})
\end{equation}
$\phi_{\mathcal{G}}$ represents the parameter in \promptG. It is assumed that the transition is deterministic and that no intermediate reward is given before the prompt is generated. With the discounted factor set to 1, the action-value function $Q_{\phi_{\mathcal{G}}}(s_t, a_t)$ equals to the state-value function of the next state $V_{\phi_{\mathcal{G}}}(s_{t+1})$. To maximize $J(\phi_{\mathcal{G}}) = V_{\phi_{\mathcal{G}}}\left(s_{1}\right)$ when $Q_{\phi_{\mathcal{G}}}(s_t, a_t) = V_{\phi_{\mathcal{G}}}(s_{t+1})$, following \cite{YuZWY17seqgan,Sutton99PolicyGradientApproximation}, the gradient of $J(\phi_{\mathcal{G}})$ can be derived as:
\small
\begin{align}
    \nabla_{\phi_{\mathcal{G}}} J(\phi_{\mathcal{G}}) 
    &= \sum_{t=1}^{n-1} \expect_{P_{1:t} \sim \promptG}\left[\sum_{a_{t}}{\nabla_{\phi_{\mathcal{G}}} \promptG(a_{t} \mid s_{t}) Q(s_{t}, a_{t})} \right]  \nonumber \\
    & \simeq  \sum_{t=1}^{n-1} \expect_{a_t \sim \pi \left( s_t, \phi_{\mathcal{G}} \right)}\left[Q(s_t, a_t)\nabla_{\phi_{\mathcal{G}}}{\operatorname{log}\promptG(a_t \mid s_t)}   \right] \label{formula:policy_gradient}
\end{align}
\normalsize
Since an initial token $p_1$ is required for the auto-regressive generator, we choose nine common initial words for $p_1$: The, It, To, There, What, This, All, If and We.
In Eq.(\ref{formula:policy_gradient}), $Q(s_t, a_t)$ is undefined, and we will discuss how to model $Q(s_t, a_t)$ to make the topic prompt task-relevant.

\paragraph{Adversarial reward.} The teacher and the student model provide delayed estimation for the generated prompts once the synthesized utterance is generated. For state $s_t = P_{1:t}$ and its corresponding action $a_t = p_{t+1}$, the content that expanded based on the temporary topic prompt is as follows:
\begin{align}
    & \hat{x} \sim \pseudoG(\cdot \mid P_{1:{t+1}}, \theta_{G}) \label{formula:extend} 
\end{align}
where $\hat{x}$ is a complete sentence on which the teacher and the student can estimate its probability for each category. Considering the probability as the likelihood that the synthesized sample belongs to a specific category, the teacher chooses the highest probability among all classes as the returned reward:
\begin{align}
     Q(s = P_{1:t}, a = p_{t+1}) = \max_{\mathcal{C}}\left(\mathcal{T}\left(\hat{x} \right)\right) \label{formula:teacher_reward}
\end{align}
where $\mathcal{C}$ is the label space. However, unitary feedback from the teacher model makes the prompt generator susceptible to overfitting to a specific prompt, which sometimes corresponds to an out-of-distribution sample (see Figure \ref{fig:ablation_prompt_example}). 
Thus, an adversarial reward is proposed, where the teacher reward is conjunct with the student reward:
\begin{align}
    & c^{\prime} = \operatorname{argmax}_{c \in \mathcal{C}} \left(\mathcal{T}\left(\hat{x} \right)\right) \nonumber \\
    & Q(s = P_{1:t}, a = p_{t+1}) = \mathcal{T}_{c^{\prime}}\left(\hat{x} \right) - \mathcal{S}_{c^{\prime}}\left(\hat{x} \right) \label{formula:reward}
\end{align}
where $c^{\prime}$ represents the category with the largest response in the teacher network. Eq.(\ref{formula:reward}) can be considered as an estimation of the discrepancy between the teacher and the student. The discrepancy is narrowed when optimizing the student, while the topic prompter is designed to enlarge this discrepancy. Once the student network has mastered an utterance guided by a certain prompt, the probability of the occurrence for this prompt should be reduced and a lower reward should be assigned. This adversarial reward allows for continuous adjustments of the prompts during the training process, significantly increasing the diversity of the pseudo-samples. 

\paragraph{Repeat penalty.} To avoid the reduction of valid information contained in the prompts due to the repetition of tokens, diversity loss is introduced to maximize the difference between the probability distributions between different tokens:
\small
\begin{equation}
    L_{repeat} = - \sum_{i=1}^n \sum_{j=1}^{i-1}\operatorname{KL}\left({\Pr(p_i|p_{< i}, \phi_{G}), \Pr(p_j|p_{< j}, \phi_{G})}\right) \label{formula:diversity}
\end{equation}
\normalsize
Algorithm \ref{algorithm:policy_prompt} describes the training procedure of \modelname. 

\begin{algorithm}[t]
  \SetAlgoLined
  \BlankLine
  \KwIn{The teacher model $\mathcal{T}$ and its parameter $\theta_{\mathcal{T}}$, the content generator \pseudoG\ and its parameter $\theta_{\mathcal{G}}$}
  \KwOut{The student model $\mathcal{S}$ and its parameter $\theta_{\mathcal{S}}$, the topic prompter \promptG\ and its parameter $\phi_{\mathcal{G}}$}
  \BlankLine
  Initial $\theta_{\mathcal{S}}$ with $\theta_{\mathcal{T}}$ \\
  \For{$i\leftarrow 1$ \KwTo $N$}{
    Fix $\theta_{\mathcal{T}}$, $\theta_{\mathcal{S}}$, $\theta_{\mathcal{G}}$ \\
    Generate $P_{1:n} = (p_1, p_2, \dots, p_n) \sim \promptG \left(\cdot ; \phi_{\mathcal{G}} \right)$ \\
    \For{$m\leftarrow 1$ \KwTo $n-1$} {
        Generate $\hat{x}$ using $P_{1:m+1}$ by Eq.(\ref{formula:extend})\\
        Compute $Q(s_m, a_m)$ by Eq.(\ref{formula:reward}) \\
    }
    Fix $\theta_{\mathcal{T}}$ and update $\theta_{\mathcal{S}}$, $\phi_{\mathcal{G}}$ \\
    Update $\theta_{\mathcal{S}}$ by Eq.(\ref{formula:KD_loss})

    Update $\phi_{\mathcal{G}}$ by Eq.(\ref{formula:policy_gradient}) and Eq.(\ref{formula:diversity})
  }
  \caption{\label{algorithm:policy_prompt} The training procedure of \modelname}
\end{algorithm}  

\begin{table}[t]
    \centering
    \resizebox{0.48\textwidth}{!}{
    \begin{tabular}{c|cc|cc|cc}
        \midrule \midrule
        & \multicolumn{2}{c|}{AG News}& \multicolumn{2}{c|}{DBPedia}& \multicolumn{2}{c}{IMDb}  \\
        & Acc. & Agree. & Acc. & Agree. & Acc. & Agree. \\
        \midrule
        \midrule
        BERT-base     & 94.75 & -    & 99.36  & -     & 88.52 & -  \\
        \midrule
        \multicolumn{7}{c}{$\text{BERT}_6$} \\
        \midrule
        Vanilla-KD    & 94.1 & 99.89 & 99.27  & 99.91 & 87.05 & 98.40 \\
        \midrule
        Random Text                           & 85.4 & 90.66 & 93.9   & 94.50 &  77.1 & 87.10  \\
        Modified-ZSKT  & 88.4 & 93.84 & FAIL   & FAIL  &  78.1 & 88.23  \\
        Modified-ZSKD     & 88.6 & 94.06 & 97.1   & 97.73 &  78.2 & 88.34  \\
        AS-DFD             & 90.4 & 95.97 & 98.2   & 98.83 &  79.8 & 90.15  \\
        \midrule
        Unlabel-KD    & 92.49 & 97.61 &  99.21 & 99.91 & 86.15 & 97.32 \\
        \modelname-Manual  & 93.29 & 98.46 &  99.10 & 99.74 & 85.83 & 96.96 \\
        \modelname-RL      & \textbf{93.74} & \textbf{98.93} & \textbf{99.30} & \textbf{99.94} & \textbf{86.94} & \textbf{98.22} \\
        \midrule
        \midrule
        
        \multicolumn{7}{c}{$\text{BERT}_4$} \\
        \midrule
        Vanilla-KD   & 93.8 & 99.57 & 99.26& 99.90 & 85.90 & 97.04\\
        \midrule
        Random Text                          & 78.5 & 83.33 & 77.3 & 77.80 & 67.6 & 76.37\\
        Modified-ZSKT & 81.1 & 96.09 & FAIL & FAIL  & 70.4 & 79.53\\
        Modified-ZSKD    & 83.8 & 88.96 & 83.0 & 83.53 & 70.7 & 79.87\\
        AS-DFD           & 88.2 & 93.63 & 94.1 & 94.71 & 77.2 & 87.21\\
        \midrule
        Unlabel-KD               & 91.54 & 96.61 & 98.97 & 99.61 & 83.86 & 94.74 \\
        \modelname-Manual             & 91.75 & 96.83 & 98.89 & 99.53 & 83.48 & 94.31  \\
        \modelname-RL                 & \textbf{92.61} & \textbf{97.74} & \textbf{99.17} & \textbf{99.81} & \textbf{85.41} & \textbf{96.49} \\
        \midrule \midrule
    \end{tabular}
    }
    \caption{Results over AG News, DBPedia and IMDB in terms of Accuracy(\%) and Agreement(\%). Since our teacher models differ from the baselines in terms of accuracy, we add a metric `Agree.' to reflect the agreement between the teacher and the student models. \label{tbl:ag_db_im_results}}
\end{table}

\section{Experiments}
\subsection{Setup}
\paragraph{Datasets.}
We conduct experiments on four text classification datasets to validate the efficacy of \modelname: AG News \cite{Zhang2015AGNews}, DBPedia \cite{auer2007dbpedia}, IMDb \cite{maas-EtAl:2011:ACL-HLT2011} and SST-2 \cite{Richard2013sst2}. 

\paragraph{Model structure.}
BERT-base is selected as the structure for the teacher model due to its widespread use. For consistency with previous methods, we use three kinds of structures for the student: $\text{BERT}_6$, $\text{BERT}_4$, and $\text{BERT}_{mini}$. As for \pseudoG, the pre-trained GPT2 \cite{radford2019language} is utilized to incorporate informative knowledge into synthesized samples, while a lightweight version of GPT2,  DistilGPT2\footnote{\url{https://huggingface.co/distilgpt2}}, is selected as \promptG\ to generate prompts that are relatively short in length.

\paragraph{Comparison methods.}
Considering that the content generator \pseudoG\ pre-trains on a large scale of unlabeled corpora, which can serve as an additional resource in the data-free setting, we include some baselines incorporating the out-of-domain data for fair comparison. Baselines are separated into three groups: data-driven knowledge distillation (Vanilla-KD), data-free knowledge distillation without any external resources (Random Text, Modified-ZSKT, Modified-ZSKD, AS-DFD \cite{Ma2020ASDFD}) and data-free knowledge distillation with out-of-domain data (Unlabel-KD and \cite{Rashid2021zskdnlp}). The introduction of each baseline is given in Appendix \ifthenelse{\equal{\AppendixFlag}{true}}{ \ref{sec:baselines}}.

\paragraph{Implementation.}
For $\text{BERT}_n$, the student uses the first n layers of the teacher network as an initialization scheme. For $\text{BERT}_{mini}$, we follow \cite{Rashid2021zskdnlp} to take the pre-trained BERT-mini to initialize the student. The maximum token length is set to 128. A grid search on parameters is performed in our experiment, where the learning rate of the student is selected from \{1e-5, 2e-5, 5e-5, 1e-4\}, the learning rate of the \promptG is selected from \{5e-6, 1e-5, 2e-5\}, the batch size is selected from \{128, 256\}, the epoch for training is set to 10, the temperature of KD is selected from \{1, 5, 10\}, and $\alpha$ is selected from \{0.5, 0.9\}. Adam is used to optimize the student network and the topic prompter. The top 50 tokens with the highest probability are selected for decoding and the threshold for top-p is set to 0.95. More details for \modelname\ and Unlabel-KD are given in Appendix \ifthenelse{\equal{\AppendixFlag}{true}}{\ref{sec:appendix-implement-manual}}.

\subsection{Experimental results}
We evaluate our proposed methods with two types of prompts: hand-crafted prompts (\textbf{\modelname-Manual}) and reinforced prompts (\textbf{\modelname-RL}). Table \ref{tbl:ag_db_im_results} and Table \ref{tbl:sst2} compare the accuracy and the teacher-student agreement of different data-free distillation algorithms. \modelname-RL outperforms all other algorithms in four datasets. A significant improvement is observed compared with AS-DFD, largely narrowing the gap between data-driven distillation and data-free distillation. Under the setting that external resources are used, \modelname\ also shows superior performance over \cite{Rashid2021zskdnlp} and Unlabel-KD. Surprisingly, in SST-2 and DBPedia with $\text{BERT}_6$ as student, \modelname-RL achieves slightly better performance compared with Vanilla-KD. This indicates that the synthesized set provides sufficient knowledge for students compared with the original training set. 

\begin{table}[t]
    \centering
    \scalebox{0.98}{
    \begin{tabular}{c|cc|cc}
        \midrule \midrule
        & \multicolumn{2}{c|}{$\text{BERT}_{6}$} & \multicolumn{2}{c}{$\text{BERT}_{mini}$} \\
        & Acc. & Agree. & Acc. & Agree. \\
        \midrule
        BERT-base                            & 93.00 & -     & 93.00 &   -  \\
        Vanilla-KD                           & 91.63 & 98.53 & 88.30 & 94.95\\
        \midrule
        $\text{Unlabel-KD}^{\diamond}$       &   -   &    -  & 84.9  & 91.29 \\
        Unlabel-KD                           & 90.94 & 97.78 & 86.24 & 92.73 \\
        \cite{Rashid2021zskdnlp}             &   -   &    -  & 85.9  & 92.37 \\
        \modelname-Manual                    & 90.48 & 97.29 & 86.35 & 92.85 \\
        \modelname-RL                        & \textbf{92.09} & \textbf{99.02} & \textbf{87.73} & \textbf{94.33} \\
        \midrule \midrule
    \end{tabular}
    }
    \caption{Performance on the dev set of SST-2. $\text{Unlabel-KD}^\diamond$ is reported in \protect\cite{Rashid2021zskdnlp}, and we re-implement Unlabel-KD with the whole set of wikitext-103.}
     \label{tbl:sst2}
\end{table}

\begin{table}[t]
    \centering
    \scalebox{0.97}{
    \begin{tabular}{c|cccccc}
        \midrule \midrule
            & 3 & 4 & 5 & 6 & 8 & 10 \\
        \midrule
        $\text{BERT}_6$ & 93.33 & 93.57 & 93.70 & \textbf{93.74} & 93.57 & 93.53 \\
        $\text{BERT}_4$ & 92.26 & 92.55 & \textbf{92.61} & 92.54 & 92.42 & 92.41 \\
        \midrule \midrule
    \end{tabular}
    }
    \caption{Accuracy comparison of the \modelname\ on AG News dataset when varying the length of the prompt. \label{tbl:prompt_length}}
\end{table}

\begin{table*}[t]
    \centering
    \scalebox{0.94}{
    \begin{tabular}{l|cc|cc|cc}
      \midrule
      \midrule
      & \multicolumn{2}{c|}{AG News}& \multicolumn{2}{c|}{IMDb} & \multicolumn{2}{c}{SST-2}\\
      &  $\text{BERT}_6$ & $\text{BERT}_4$ & $\text{BERT}_6$ & $\text{BERT}_4$ & $\text{BERT}_6$ & $\text{BERT}_{mini}$ \\
      \midrule
      \modelname-Manual                 & 93.29 & 91.75 & 85.82 & 83.48 & 89.11 & 78.55\\
      \modelname-RL                     & 93.74 & 92.61 & 86.94 & 85.41 & 92.09 & 87.73\\
      \midrule
      w/o Adv Reward                & 93.57(0.17$\downarrow$) & 92.47(0.14$\downarrow$) & 86.75(0.19$\downarrow$) & 84.87(0.54$\downarrow$) & 88.42(3.67$\downarrow$) & 81.08(6.65$\downarrow$)\\
      w/o Repeat Penalty        & 93.01(0.73$\downarrow$) & 92.45(0.16$\downarrow$) & 86.69(0.25$\downarrow$) & 84.67(0.74$\downarrow$) & 91.97(0.12$\downarrow$) & 86.70(1.03$\downarrow$)\\
      w/o Adv Reward + Repeat Penalty  & 92.62(1.12$\downarrow$) & 92.36(0.25$\downarrow$) & 86.58(0.36$\downarrow$) & 84.90(0.51$\downarrow$) & 88.30(3.79$\downarrow$) & 79.01(8.72$\downarrow$)\\
      \midrule
      \midrule
      \end{tabular}
      }
      \caption{Ablation results for \modelname. ``w/o Adv Reward" denotes the adversarial feedback is replaced by Eq.(\ref{formula:teacher_reward}). ``w/o Repeat Penalty" indicates the objective Eq.(\ref{formula:diversity}) is not included in the training objective. \label{tbl:ablation}}
\end{table*}

\subsection{Quantitative Analysis}

\begin{figure*}[t]
    \centering
    \scalebox{0.42}{
    \includegraphics{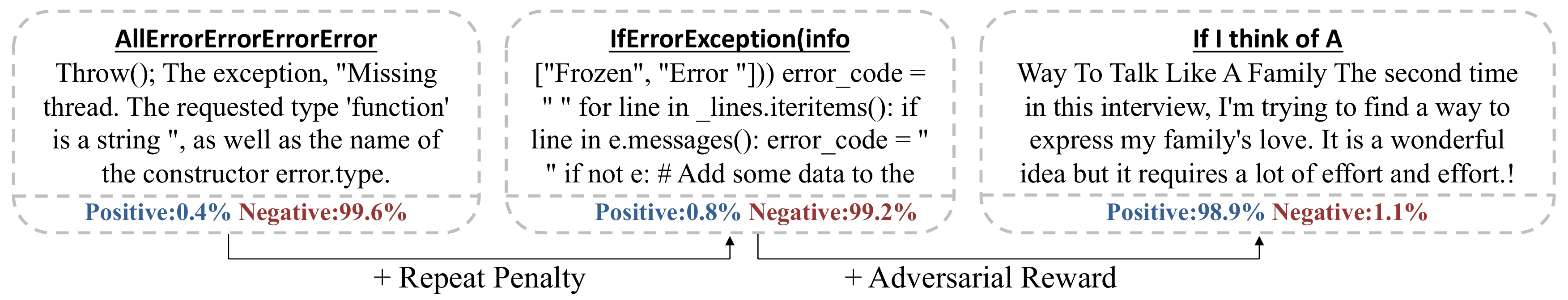}
    }
    \caption{Generated prompt with different modules in \modelname. This prompt is optimized under the SST-2 dataset.}
    \label{fig:ablation_prompt_example}
    \vspace{-2mm}
\end{figure*}

\begin{figure}[t]
    \centering
    \scalebox{0.42}{
    \includegraphics{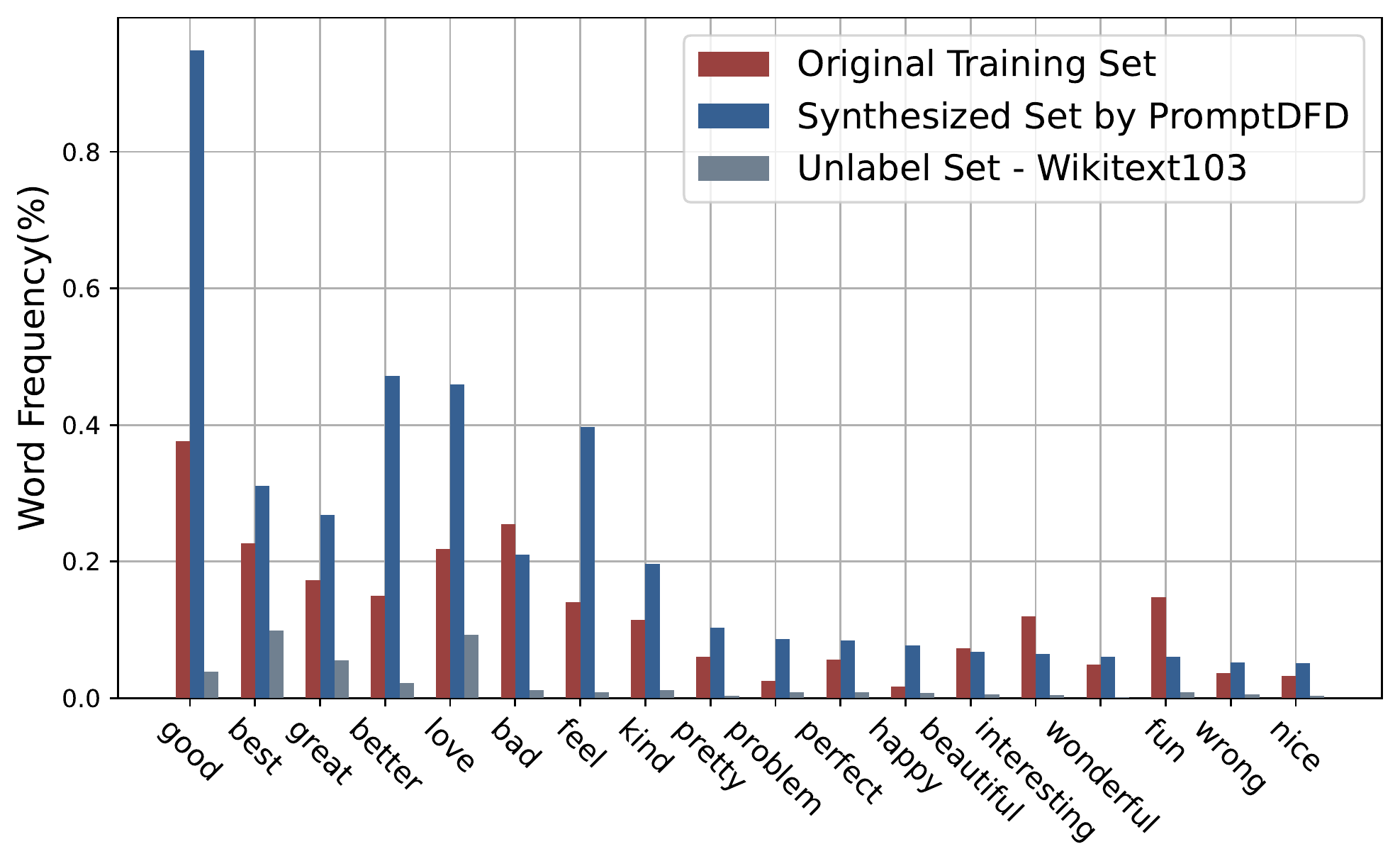}
    }\vspace{-1mm}
    \caption{Word frequency of some positive/negative words in the training set of SST-2, synthesized samples generated by \modelname\ and wikitext103. }\vspace{-2mm}
    \label{fig:word_frequency}
\end{figure}

\paragraph{Impact of prompt length.} Table \ref{tbl:prompt_length} shows how the length of the prompt influences the distillation performance. The case where the prompt length is extended until it reaches the length of the pseudo sample can be characterized as a direct fine-tuning of the synthesized sentence. According to the table, the optimal length for prompts is approximately 5–6 words. If the prompt is too short, there will be insufficient prefix constraints in the prompt, while lengthening the prompt results in a decrease in performance due to the increasing complexity of training \promptG\ with a huge action space.

\paragraph{Ablation study.} Ablation results are reported in Table \ref{tbl:ablation}. Compared with \modelname-Manual (designed with templates in table \ref{tbl:manual_prompt}), the reinforced prompts can steadily improve the performance, indicating the hand-crafted prompts are not the best choice for pseudo-sample construction. The repeat penalty reduces the redundancy between tokens, showing a 0.12\%-0.74\% decrease on the performance. Adversarial reward shows a little effect on AG News and IMDb but observes a significant impact on SST-2. The main reason for this is that the AG News and IMDb datasets are relatively simpler than SST-2, which can be demonstrated by the performance of \modelname-Manual.

An example of how each module affects the generation of prompts is shown in Figure \ref{fig:ablation_prompt_example}. The original prompt is quite repetitive, which is mitigated by introducing a diversity loss. Nevertheless, we find that the generated prompts are far from the domain of the dataset. The teachers give a high likelihood to OOD samples, but they are not helpful in distilling, which is reflected in the performance of models that are distilled using these samples. While the adversarial reward does not prevent the model from generating such OOD samples, a near-zero reward is given after the student has learned these samples. The topic prompter is biased not to generate those prompts, avoiding the prompt generators from getting stuck in out-of-distribution samples.

\begin{table}[t]
    \centering
    \begin{tabular}{c|cc|cc}
        \midrule \midrule
         & \multicolumn{2}{c|}{IMDb} & \multicolumn{2}{c}{SST-2} \\
         & $\text{BERT}_{6}$ & $\text{BERT}_{4}$& $\text{BERT}_{6}$ & $\text{BERT}_{mini}$\\
         \midrule
        \modelname\   & 86.94 & 85.41 & 92.09 &  87.16  \\
        w/o semantic  & 84.99 & 83.17 & 90.25 &  85.44  \\
        \midrule \midrule
    \end{tabular}
    \caption{Accuracy when the semantics of the synthesized utterances are broken down by shuffling the words.}
    \vspace{-2mm}
    \label{tbl:shuffle}
\end{table}

The quality of the synthesized samples are evaluated by two criteria: the presence of task-related terms and the reasonableness and meaningfulness of the utterance.

\paragraph{Topic relevance of synthesized samples.}
To visually demonstrate the relationship between the synthesized samples and the original training data, we analyze the frequency of task-related keywords in Figure \ref{fig:word_frequency}. On the sentiment classification dataset SST-2, \modelname\ tends to generate more emotional words than the unlabeled corpus like wikitext103. These words frequently appear in the original training set, demonstrating that the synthesized set generated by \modelname\ mimics the original data to some extent.

\paragraph{Importance of synthesizing meaningful samples.}
To destroy the semantics in the synthesized samples generated by \modelname-RL, we shuffle the words, resulting in a synthesized set that shares the same set of words with broken meanings. Taking these disordered sentences as the distillation set, a noticeable performance drop is observed in Table \ref{tbl:shuffle}, showing that plausible synthetic utterances are indispensable in data-free knowledge distillation.

\section{Conclusion}
In this work, we propose a novel data-free knowledge distillation framework \modelname\ that incorporates controllable language prior into the synthesized samples by reinforced prompts. The reinforced prompt is guided by the adversarial feedback to adjust the topic dynamically and increase the diversity. \modelname\ largely improves the quality of the synthesized samples and the distillation performance, and even reaches the performance of data-driven knowledge distillation with the original training data. 

\section*{Acknowledgments}
This work is supported by the Key Research and Development Program of Zhejiang Province, China (No. 2021C01013), the National Key Research and Development Project of China (No. 2018AAA0101900), CKCEST, and MOE Engineering Research Center of Digital Library. This work is also supported by NUS Faculty Research Committee Grant (WBS: A-0009440-00-00) and Advanced Research and Technology Innovation Centre (Project Reference: ECT-RP2).

\clearpage
\bibliographystyle{named}
\bibliography{ijcai22_simplify}

\clearpage
\appendix

\section{Appendix}

\subsection{Baselines} \label{sec:baselines}
We introduce the baselines used in the experiments here. As mentioned before, the methods are divided into three types, depending on the resources they use in DFKD.
\begin{itemize}
    \item \textbf{Vanilla-KD}\cite{Hinton2015Distill}. Using the original training data to train the student with the soft logits from the teacher and the hard target from human-annotated labels.
    \item \textbf{Random Text}. A baseline method mentioned in \cite{Ma2020ASDFD}, which randomly selects words from the  vocabulary released with the pre-trained model.
    \item \textbf{Modified-ZSKT, Modified-ZSKD}. Baselines that mentioned in \cite{Ma2020ASDFD}, which adapts the DFKD algorithms in computer vision \cite{micaelli2019zerozskt,nayak2019zerozskd} to text classification tasks.
    \item \textbf{AS-DFD}\cite{Ma2020ASDFD}. The first DFKD algorithm designed for text. AS-DFD proposes to construct pseudo embeddings and designs an adversarial self-supervised module to encourage diversity.
    \item \textbf{Unlabel-KD}. Knowledge distillation that takes wikitext-103 \cite{MerityXBS16} as the out-of-domain corpora. Softened logits are used to match the student model with the teacher model. Implementation details are provided in Appendix \ref{sec:appendix-implement-unlabel}.
    \item \textbf{\cite{Rashid2021zskdnlp}} combines out-of-domain data and model inversion to train a generator.
\end{itemize}

\subsection{Implementation Details}
\subsubsection{Unlabel-KD} \label{sec:appendix-implement-unlabel}
We adopted a similar approach with Vanilla-KD, in which the original dataset is replaced by wikitext-103. Wikitext-103 is preprocessed (including filtering excessively short statements, cleaning some special formats in wikitext-103, etc.) and is used as the transfer set in knowledge distillation to train the student network. In the experiments, the number of training epochs is set to 2 and the learning rate is searched from \{1e-5, 2e-5, 5e-5\}. A scheduler for the learning rate is adopted in the training process: the first 10\% steps are the start-up steps with the learning rate increasing uniformly, and the learning rate decreases gradually to 0 in the next 90\% steps. In addition, the weight decay is selected from \{1e-5, 1e-8\} and the temperature parameter of distillation is selected from \{1, 5, 10\}.

\subsubsection{\modelname-Manual} \label{sec:appendix-implement-manual}
As for the manually-designed prompt, the designed template for each dataset is shown in Table  \ref{tbl:manual_prompt}. The category names will be randomly filled into the placeholder to form a complete prompt. \modelname-Manual takes the same initialization way as \modelname-RL. The optimal combination of parameters for \modelname-Manual is selected from: \{1e-5, 2e-5, 5e-5, 1e-4\} for the learning rate of the student network , \{128, 256\} for the batch-size, \{1, 5, 10\} for the temperature of knowledge distillation, \{0.5, 0.8\} for $\alpha$. As for the remaining hyper-parameters, \modelname-Manual follows the setting in \modelname-RL.

\subsection{Examples for \modelname-RL and \modelname-Manual} \label{sec:appendix-case-study}

\begin{table*}[!ht]
    \centering
    \begin{tabular}{cp{13.6cm}c}
    \midrule \midrule
    \makecell{Dataset-\\Category}  & \centering Synthesized Sample & \makecell{Teacher-\\Confidence} \\
    \midrule
    \multirow{1}*{\makecell{AG News-\\Science}} & \underline{A latest science news} item Scientists are finding a way to send high-energy cosmic rays to Earth's surface. The first time they could see them was with a pair of telescopes. The phenomenon was first spotted by the German research team known as the Minkiwi Observatory in Germany. Minkiwi, commonly known as the Dark Eye telescope, is part of the Gaia constellation, or Earth-wide space telescope and is now orbiting Earth. The telescope was launched from Roswell, New Mexico, in 1994. & 0.998\\
    \midrule
    \multirow{1}*{\makecell{DBPedia-\\Album}} & \underline{A document about an album} we're listening to has been posted to the fan site Reddit. To add to the noise I'm sure, I'm not the first. I mentioned that I found this ``universe" pretty awesome, but that was just part of the fun. We are always looking for new things for our projects, and this was another example of our love for these kinds of things. We have a long way to go, but we have a lot of things we can make with the internet that we like, so if we are able to come up with new things that can make things for a new user & 0.644 \\
    \midrule
    \multirow{1}*{\makecell{IMDb-\\Negative\\Review}} & \underline{A negative movie review} has already emerged on the internet, showing that the actors did not really believe that they were being followed and that the people watching them were not the ones who were doing the real things. The critics are taking issue with this in an article titled `The Movie Is Lying.' The movie itself is not really a documentary, but a book that a writer found on his own. The book, entitled `The Movie Is Lying,' is a book written by a man in the early 1950s who was known as Arthur Schreier, who was known as Arthur Schreier.& 0.740 \\ 
    \midrule
    \multirow{1}*{\makecell{SST-2-\\Positive}} & \underline{positive sentence:} ``The entire thing was a success and we're still going to be around to see if it works." So it sounds like this is an important milestone. If we don't make changes in this direction, what happens to the world? & 0.993\\
    \midrule \midrule
    \end{tabular} 
    \caption{Examples generated by \modelname-Manual} \label{tbl:case_study_manual}
\end{table*} 

\begin{table*}[htbp!]
    \centering
    \begin{tabular}{cp{13.6cm}c}
    \midrule \midrule
    \makecell{Dataset-\\Category}  & \centering Synthesized Sample & \makecell{Teacher-\\Confidence} \\
    \midrule
    \multirow{1}*{\makecell{AG News-\\World}} & \underline{The President Donald Trump} and First Lady Melania Trump were at their most recent events together after an early afternoon meeting with First Lady Melania Trump at the White House. Trump took the unusual step of meeting with First Lady Melania Trump while sitting next to the president in the Oval Office. Just over an hour later, a private meeting was had in progress. President Donald Trump framed the meeting as ``a very friendly one" and said no one should be ``sad" if he had made ``many more bad moves," the New York Daily News reported on Oct. 22.  & 0.825\\
    \midrule
    \multirow{1}*{\makecell{DBPedia-\\Company}} & \underline{They Wright Doug Bubble} and John McNeill, who are married in Iowa, went to an auto plant that has a big industrial plant under construction. One of them was built, the other one closed. ``It's a tremendous loss for this town," said Jeff Dolan, who lives in Hawkeye State. But it's not a loss that should discourage people from going to this place: The area is home to thousands of auto makers and the state has some of the largest auto dealerships in the country. This year, more than 200 companies made a total of nearly \$6 billion in profit at the factory that had a major & 0.942 \\
    \midrule
    \multirow{1}*{\makecell{IMDb-\\Positive\\Review}} &\underline{The Jun music is fantastic} and very similar to Kinks Music. Music video for Kinkz is a very good piece but a little over 90\% of it is not perfect or funny. This was the case when I wrote this post. My wife had to use my laptop after school (not to mention my mom's computer in the morning), so I would have had to switch to Kinkmusic at the end of the day. Then I used it and realized that no one could take it. I wanted my own music to be as fun as the music videos I'd seen of it.& 0.944\\ 
    \midrule 
    \multirow{1}*{\makecell{SST-2-\\Positive}} & \underline{If I Think Of A} Way To Talk Like A Family The second time in this interview, I'm trying to find a way to express my family's love. It is a wonderful idea but it requires a lot of effort and effort. I want to spend time talking with you on the phone, and we're going to do this on every phone. It would take a lot of effort to find that perfect moment and that perfect feeling to talk about your feelings, and that one moment you might never give up the other. That's going to be hard for a lot of people. But it's going to be true intimacy.& 0.989\\ 
    \midrule \midrule
    \end{tabular} 
    \caption{Examples generated by \modelname-RL} \label{tbl:case_study_rl}
\end{table*} 

Table \ref{tbl:case_study_manual} and Table \ref{tbl:case_study_rl} show some examples in AGNews, DBPedia, IMDb and SST-2. For \modelname-RL, we observe some interesting phenomena, where \modelname\ can be seen as interpretability for the model.

\paragraph{AG News}
The utterances generated for the AG News are evidence that the reinforced prompts and the synthesized samples are closely related to the setting of the AG News. Compared to the hand-crafted prompts, these prompts are more effective and introduce more information into the generative process of the GPT2.

\paragraph{DBPedia}
As the DBPedia dataset is simple, the topic prompter has fewer restrictions. As the prompter nears its final phase of training, a large number of names of people and places appear in the prompts, which correspond to the categories of athletes, artists, buildings, places, and villages in the dataset. 

\paragraph{IMDb}
The example for IMDb reveals that the keyword ``movie review" in the label is not essential for this dataset, and that it is the emotional content of the review that dominates the teacher model. Although the example is evaluating a song, it contains a great deal of subjective assessment, much like a movie review, and the teacher model gives a conditional probability of 0.944 that the sample will be evaluated as a positive review.

\paragraph{SST-2}
A keyword `think' appears in the generated prompts. This keyword directly leads the subsequent content to be an evaluative paragraph, which is generally emotional. It also suggests that we can use some neutral words or subject words when designing the prompts manually.
\end{document}